\documentclass[runningheads]{llncs}

\usepackage{hyperref}
\usepackage[numbers]{natbib}
\usepackage{amsmath}
\usepackage{amssymb}
\usepackage{times}
\usepackage{bm}
\usepackage{natbib}
\usepackage{inputenc}
\usepackage{amsmath}
\usepackage{graphicx,psfrag,epsf}
\usepackage{enumerate}
\usepackage{natbib}
\usepackage{url} 
\usepackage{xcolor}
\usepackage{multirow}
\usepackage{graphicx}
\usepackage{todonotes} 

\begin{document}
	\title{The Thousand Faces of Explainable AI Along the Machine Learning Life Cycle: Industrial Reality and Current State of Research}
\titlerunning{The Thousand Faces of Explainable AI Along the ML Life Cycle}

\author{Thomas Decker \inst{1,2}\and
	Ralf Gross\inst{1} \and
	Alexander Koebler\inst{1,3} \and
	Michael Lebacher\inst{1} \and
	Ronald Schnitzer\inst{1,4} \and
	Stefan H. Weber\inst{1} \thanks{Equal contributions with alphabetical order.}
}

\authorrunning{T. Decker et al.}

\institute{Siemens AG, Munich, Germany \and
	Ludwig Maximilians Universität Munich, Munich, Germany \and
	Goethe University Frankfurt, Frankfurt, Germany \and
	Technical University of Munich, Munich, Germany\\
	\email{\{thomas.decker; ralf.gross; alexander.koebler; michael.lebacher; ronald.schnitzer; stefan\_hagen.weber\}@siemens.com}}

\maketitle

\begin{abstract}
In this paper, we investigate the practical relevance of explainable
artificial intelligence (XAI) with a special focus on the producing industries and
relate them to the current state of academic XAI research. Our findings are
based on an extensive series of interviews regarding the role and applicability of
XAI along the Machine Learning (ML) lifecycle in current industrial practice
and its expected relevance in the future. The interviews were conducted among
a great variety of roles and key stakeholders from different industry sectors. 
On top of that, we outline the state of XAI research by providing a concise review of the relevant literature. This enables us to provide an encompassing overview covering the opinions of the surveyed persons as well as the current state of academic research. 
By comparing our interview results with the current research approaches we reveal
several discrepancies. While a multitude of different XAI approaches exists, most
of them are centered around the model evaluation phase and data scientists. Their
versatile capabilities for other stages are currently either not sufficiently explored
or not popular among practitioners. In line with existing work, our findings also
confirm that more efforts are needed to enable also non-expert users’ interpretation
and understanding of opaque AI models with existing methods and frameworks.

		
\end{abstract}

\section{Introduction} \label{intro}
Artificial Intelligence (AI) has become increasingly pervasive in the industry and proved to be successful in multiple applied industrial use cases \citep{wang2018deep, wuest2016machine, meng_machine_2020}. However, it is still a challenging step from providing first Proof of Concepts (PoCs) to actually deployed Machine Learning systems (e.g., \cite{sculley2015hidden}). While the causes for this problem are manifold, one potential reason is the notorious black-box nature of AI, which prevents AI developers from understanding and communicating their models, hampers the trust-building process, impedes efficient communication with stakeholders, and complicates monitoring and maintenance. Therefore, the problem of opaque AI poses challenges along the entire AI life cycle \citep{studer2021towards}.

\begin{figure}
	\includegraphics[trim={0 11cm 0 7.4cm},clip,width=\textwidth]{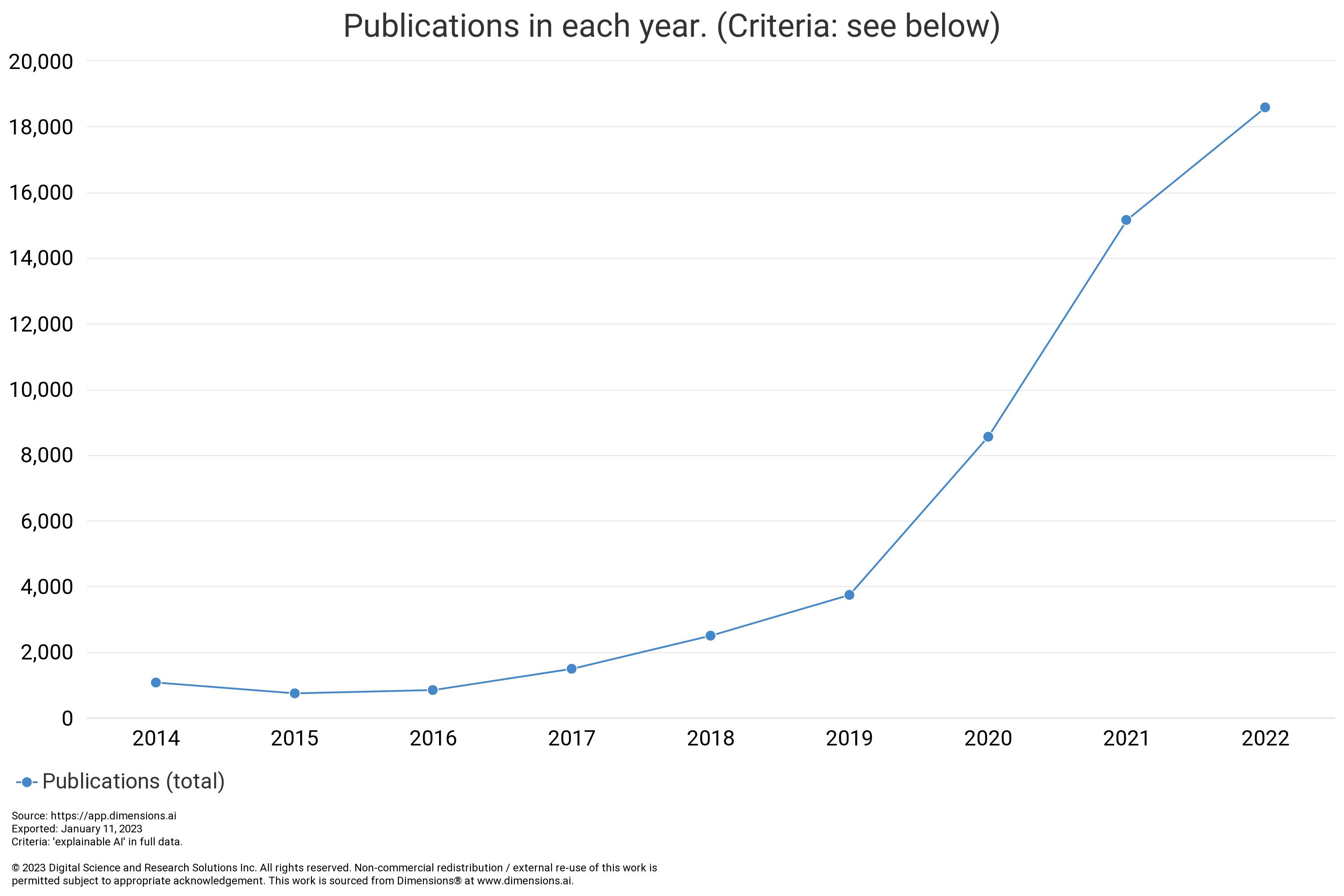}\label{XAItrend}
	\caption{Number of publications related to 'Explainable AI' from 2014 to 2022. Source: \href{https://app.dimensions.ai/}{dimensional.ai}, accessed 2023/20/01}
\end{figure}

For this reason, explainable Artificial intelligence (XAI) has established itself as a multifaceted research field covering a vast variety of approaches and incorporating perspectives from different academic areas.
We define explainability, in accordance with \cite{doshi2017towards} as {\sl "any technique that provides the ability to explain or present the outcomes or predictions of AI systems in understandable terms to humans"}. The relevance of such techniques is mirrored by the fast-growing interest in academia and the industry's demand. One manifestation is the number of research papers published in the area of XAI as shown in Figure \ref{XAItrend}, see also \cite{adadi2018peeking}. 
Given this rapid growth in research interest, the field has already accumulated an enormous amount of methods and tools for models applied to image data \cite{simonyan2013deep}, text data \cite{galassi2020attention, ribeiro2016should}, tabular data \cite{lundberg2017unified, guidotti2018local}, time series data \cite{rojat2021explainable} and reinforcement learning \cite{wells2021explainable}. For a more encompassing view on the topic, we refer to \cite{molnar2020interpretable, doshi2017towards}. 

However, most current academic research effort is still directed towards researchers and AI developers and does not focus on other end users or the role of explainability along the whole machine learning life cycle. Notable exceptions are the studies by \cite{bhatt2020explainable}, and \cite{dhanorkar2021needs}  that concern the role of XAI for deployment and across the ML life cycle, respectively. The researchers in \cite{bhatt2020explainable} conducted 50 interviews in domains such as Finance, Insurance, and Content Moderation, finding that ML experts and developers increasingly use XAI techniques for error tracing and debugging. They highlight that XAI currently bears no benefit to other stakeholders, and a general gap exists between the potential usage scenarios and the actual practice of XAI. In \cite{dhanorkar2021needs}, 30 interviews have been conducted with a focus on natural language processing (NLP) researchers, pointing on the finding that XAI rarely addresses challenges along the full AI life cycle  in practice.

We aim to extend and supplement these findings by providing a real-world perspective from the industry with a comprehensive coverage of the role of XAI in the ML life cycle. Based on a qualitative analysis of semi-structured interviews, we integrate the different views and perspectives on this topic in a consistent picture and set our findings in the context of the current literature. This allows us to highlight challenges not yet addressed and research gaps. Our contributions to the literature are as follows:
\begin{itemize}
	\item We conducted 36 semi-structured interviews with practitioners and various stakeholders to identify the current relevance of XAI along the ML life cycle with a focus on the producing industries.
	\item We surveyed current XAI research and allocated it to the best matching ML life cycle stage.
	\item We juxtaposed and compared both perspectives to identify alignments and mismatches.
\end{itemize} 
It is clear that this paper cannot summarize the findings of the enourmous XAI literature as a whole and is, therefore, restricted to the research works we found most fitting in the context of our interviews and the respective life cycle stages.

In the remainder of the paper, we will provide background on the interviews and methodology in section \ref{background}. This is followed by section \ref{motivation}, presenting a broad perspective on how the need for XAI is motivated in academia, in applied industrial use cases, and how this topic relates to regulations. Then, in section \ref{MLCycle}, we shed a light on the relevance of XAI along the ML life cycle. We end the paper with a summary on our reserarch hypothesis in section \ref{hypotheses_summary} and a conclusion in section \ref{conclusion}. 

\section{Background on the Interviews and Methodology}\label{background}
We conducted 36 remote interviews ranging from 45 to 60 minutes with employees from Siemens AG, startups, technical associations, and research institutes. The main focus, with 19 persons in the sample, is on the role of {\sl data scientists}, defined here as well-trained but applied working persons that solve practical problems with machine learning in the domain of industrial automation and autonomous vehicles. The focus on the data scientists stems from the fact that this role usually needs to provide support at all stages of the ML life cycle.  However, we also integrate the views from (applied) academia with three {\sl machine learning researchers} from public research institutes. Furthermore, we interviewed two {\sl certification and standardization engineers} from public institutions and two {\sl safety engineers} to integrate their views. In order to cover the organizational and business perspective, we also interviewed {\sl machine learning team leads}, {\sl sales persons}, {\sl machine learning product and project managers}, and even two {\sl chief technology officers}. These persons are subsumed in the role of {\sl managers}. Lastly, we interviewed two {\sl machine learning and IT service technicians} as well as one {\sl domain expert} working in close collaboration with data scientists. Hence, we can roughly separate two groups, a technically oriented group and a group of team leads and managers. See Table \ref{interview_overview} for a comprehensive overview of the covered roles, expertises, and domains.
\begin{table}[]\small \label{interview_overview}
	\caption{Number of Interviews by domains (columns) and roles (rows)}
	\begin{tabular}{lcccccc} 
		Role/Domain & Industrial  &Technical   & Autonomous   & Startups  & Research & $\sum$ \\
		&Automatization&Associations&Vehicles&&Institutes\\ \hline 
		Data Scientist	& 7 & - & 1 & 3& - & 11 \\
		ML Researcher 	& - & - & - & - &3 &  3 \\
		Sales & 3&- &-&-&-&3\\ 
		Safety Engineer & 2 &-&1& -&-&3 \\
		ML Team Lead & 1 & -&-&2&- &3 \\
		ML Service Technician & 2&-&-&-&-&2 \\
		Certification Engineer&-&2&-&-&-&2\\
		Manager &3&-&3&2&-&7\\
		Domain Expert &1&-&-&-&-&1\\ \hline \hline
			$\sum$ &19&2&5&7&3&36\\ \hline \hline
	\end{tabular}
	
\end{table}

We ensured all our interview partner anonymity. Hence, except for Siemens AG, we do not list companies or names of interview partners, and we refrain from direct quotes. Similar to \cite{holstein2019improving}, \cite{dhanorkar2021needs} and \cite{bhatt2020explainable}, we rely on semi-structured interviews that are guided by underlying hypotheses for the different roles under study. See Table \ref{hypotheses} for our main guiding hypotheses for data scientists and machine learning researchers (first four rows). We further investigated the hypotheses in rows five to eight for interview partners responsible for monitoring and maintaining AI systems. Finally, the hypotheses for the more business-oriented group are shown in the last three rows. The formulation of the hypotheses was guided by the idea of covering multiple stages of the Machine Learning life cycle as well as typical tasks that are presumably relatable to XAI.

\begin{figure}
	\includegraphics[trim={0cm 0cm 0cm 0 0cm}, clip=True, width=1.4\textwidth, angle=90]{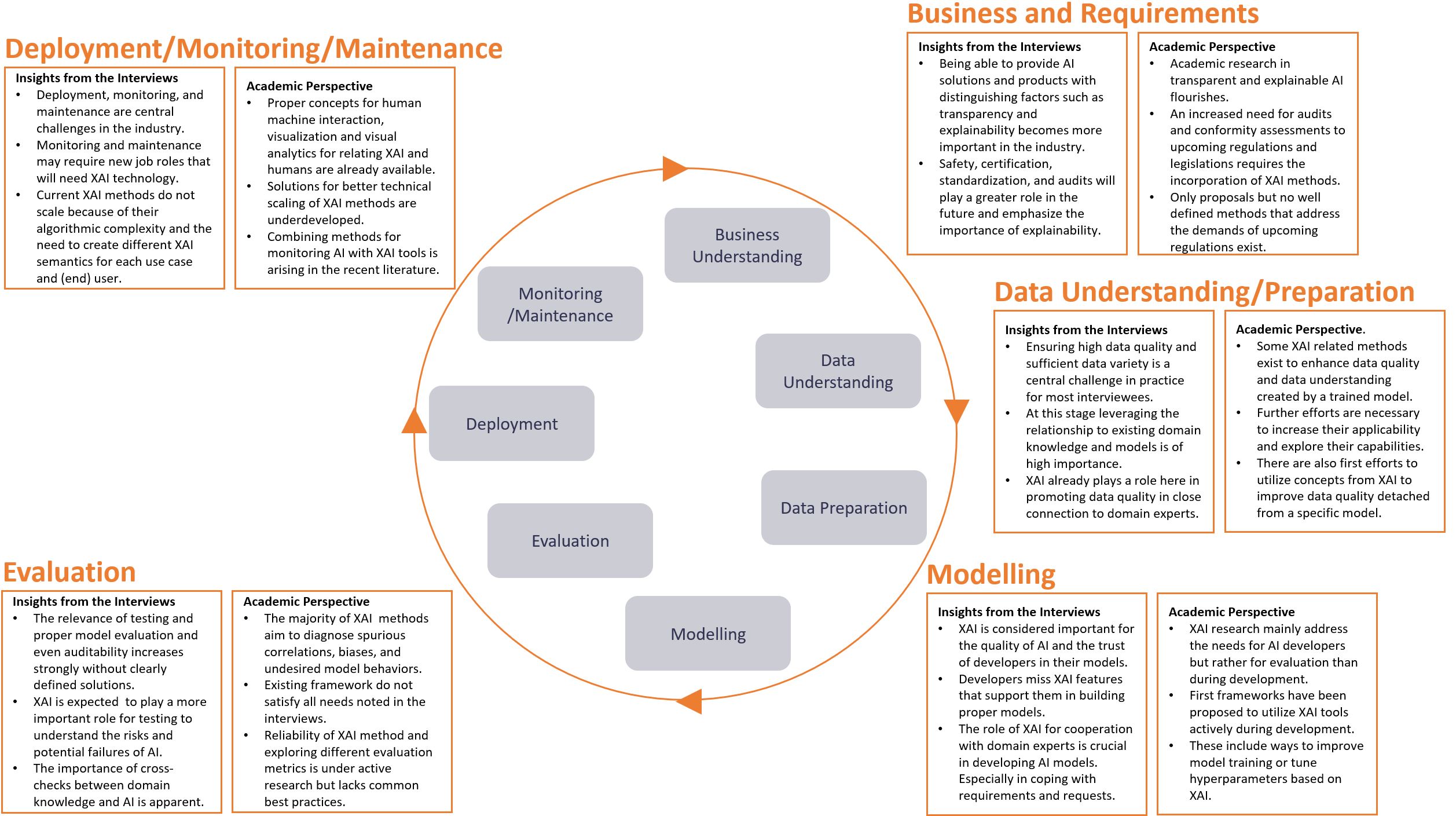}
	\caption{CRISP-ML life cycle with summarized findings. For each stage, we display insights from the interviews contrasted with the academic perspective.}\label{crisp}
\end{figure}

Where appropriate, we also asked which XAI tools are currently used and which obstacles concerning the scaling of XAI have been encountered. Based on the transcribed protocols, we extracted main insights and matched them along the ML life cycle based on CRISP-ML \cite{studer2021towards}. This life cycle model builds upon the CRISP-DM model, first introduced in 2000 \cite{wirth2000crisp}. Although being relatively old, the CRISP-DM process is still considered the most widely used analytical methodology for data mining and knowledge discovery projects \cite{martinez2019crisp}.
CRISP-ML can be interpreted as an adoption of the CRISP-DM model towards the particular requirements of machine learning applications, especially concerning the full coverage of the whole life cycle. It divides the life cycle into six stages, which are:
\begin{enumerate}
	\item
	\textit{Business and Data Understanding}, covering the scoping of ML applications, including building success criteria and feasibility concerns, as well as collecting and verifying the quality of the data.  
	\item
	\textit{Data Preparation}, considering all necessary data preparation steps, such as selecting, cleaning, and standardizing the data.
	\item
	\textit{Modelling}, addressing any required step for bringing up the model, including model selection, training, and potentially pruning.
	\item 
	\textit{Evaluation}, concerning the validation of the model performance and having in mind deficiencies of the model such as lack of robustness as well as the success criteria defined in the first stage. On top of that, explainability for AI practitioneers and end users is explicitly mentioned.
	\item 
	\textit{Deployment}, addressing the implementation of the model into the appropriate hardware and validating the model again under production conditions.
	\item 
	\textit{Monitoring and Maintenance}, considering possible changes in the environment or the application itself, possibly influencing the model performance. Thus, it is required to monitor the model and potentially adapt it to changes in production conditions.
\end{enumerate} 
For a more detailed explanation of the individual stages of the CRISM-ML model, we refer to \cite{studer2021towards}. Figure \ref{crisp} provides a summary of our main findings along the ML life cycle.

Apart from the life cycle we also included the general motivation for using XAI as a focust category, covering business aspects and the role of XAI in regulation, standardization, and safety aspects (see section \ref{motivation}). The reason for this additional section is that, during the interviews, we learned that these aspects are important as a general motivation for XAI in real-world applications but cannot be directly allocated to the ML life cycle.  

\begin{table}[]\small \label{hypotheses} 
	\caption{Guiding hypotheses for the interviews related to different roles}
	\begin{tabular}{l|l} \hline \hline
		Hypotheses Data Scientists & XAI support the communication with domain experts.                  \\
		& XAI improves the development process.                                 \\
		& XAI improves AI testing.                                             \\
		& XAI relives from the lack of trust in the developed models.          \\ \hline
		Hypotheses Monitoring                       & XAI supports the task of monitoring.                                 \\
		& XAI can support the task of maintaining AI.                                \\
		& XAI can support root cause analysis, commissioning, and other tasks. \\
		& XAI can support audits.                                                  \\ \hline
		Hypotheses Business                         & XAI and AI are among the strategic priorities.                        \\
		& XAI bridges gaps in cross-functional teams.                          \\
		& XAI is needed as a distinguishing factor (from competitors).        \\ \hline \hline
	\end{tabular} 
	
\end{table}

\section{Motivation for Explainability} \label{motivation}

\subsection{Insights from the Interviews}
\subsubsection{The Need for Distinguishing Factors}\label{business}
According to our interviews, companies that try to sell AI products or solutions are increasingly confronted with their customers' growing AI maturity, which changed their expectations towards AI products and solutions. Consequently, competition became more fiercely and the pressure to provide distinguishing factors increased in order to match the demands of customers and their expectations regarding the extent of AI offering packages. It is important, however, to note that the demand for distinguishing factors has the precondition that customers have already achieved a certain degree of AI maturity, and the placement of distinguishing factors can also be a matter of market timing. Among the most important distinguishing factors are {\sl transparency, traceability} and {\sl explainability}, that promise to improve trust and acceptance of AI solutions and products. Forerunners for setting this trend are companies such as Google, Meta, Amazon, and Microsoft, which started activities to provide explainability, robust AI, trustworthy AI offerings, and software.\footnote{Examples are, e.g., the explainable AI frameworks and tools by Google for the \href{https://cloud.google.com/explainable-ai}{Google Cloud}, the Captum library \cite{kokhlikyan2020captum} by Meta/Facebook, \href{https://aws.amazon.com/de/lookout-for-equipment/}{Amazon Lookout} as well as \href{https://aws.amazon.com/sagemaker/clarify/?nc1=h_ls&sagemaker-data-wrangler-whats-new.sort-by=item.additionalFields.postDateTime&sagemaker-data-wrangler-whats-new.sort-order=desc}{Amazon SageMaker Clarify} Model Explainability and Microsofts InterpretML \cite{nori2019interpretml} python library  (Comments of the authors and not part of the interview responses). }  

Although transparency and explainability are important factors, we also found that trust in the data business is not only of technical nature but also related to general factors such as business habits, trust in customer relations, and brand perception. Even companies with well-established and trustworthy brands can struggle with mistrust concerning their AI business models. 

\subsubsection{Standards, Regulation and Safety}\label{Safety}
Besides the need for distinguishing factors, the topic of upcoming regulations and standards can also drive the need for XAI. Although these topics were of high interest to many interviewees, their views on future standards were in conflict. However, common ground was that currently, almost no, only bad, or too lax external (and internal) standards are in place and that the future will hold standards, regulations, and even regular audits for AI.

Regarding the expected scope of formal AI certification, we learned that safety-critical domains, such as {\sl healthcare, public transport, public services, industrial automation} and {\sl shopfloor control} are likely to be subject to certification in the future.  

Looking at the certification methodology, some persons advocated black box testing, e.g., with pre-defined and hidden datasets available only to certification authorities. This is also in line with the argumentation that certifying units do not need to understand (X)AI and any details besides functionality, performance or statistical arguments do not matter to them. In contrast, other interviewed persons argued that certification authorities will assume and demand that the developers understand their models in depth - which gives a strong pointer towards explainability. This explains why using XAI for certification was promoted by some interviewees. One approach would be that certification authorities use XAI methods to understand the AI to be certified, meaning that XAI will become an own component in the certification process. Another approach is to place the explainability on the side of the developers, where employment of XAI methodology is a mandatory feature to satisfy regulatory requirements. 

Focusing on the more specific area of AI safety, we found that it is hard to evaluate AI systems with classical safety approaches (see also Section \ref{development}). The main issue is the fact that AI suffers from the {\sl correlation versus causation problem} while safety arguments typically rely on causation. XAI was credited with providing a valuable set of tools for identifying errors, safety verification, and support for safety audits. Regarding the limitations of XAI for safety argumentations, we learned that XAI is currently not known to relevant stakeholders such as safety engineers and cannot guarantee the safety or achieve a safety claim on its own. Hence, in the future, it is likely that XAI tooling will not support safety argumentations solely but in combination with other approaches.

\subsection{Academic Perspective on Regulation and XAI}

\subsubsection{Current Proposals for Regulation}
Essentially, we learned from the interviews that conflicting views and even confusion concerning standards, regulations, and safety requirements for AI were predominant among the interviewed persons. This was somewhat surprising as there already exist quite concrete proposals.
The European Commission (EC) has presented a draft for the legislation and regulation of Artificial Intelligence \citep{AIAct} (AIA). Similar approaches exist in other parts of the world, such as the United States. For example, the Algorithmic Accountability Act of 2022 (AAA US) \cite{AAA_US}, which is compared to the AIA in the literature, see \cite{mokander2022us,gstrein2022european}.
Even though compared to the AIA, the AAA US is less concrete and ambitioned \cite{gstrein2022european}, it is expected that the European AI regulations will implicitly expand globally.
This phenomenon, also called \emph{de facto Brussels Effect}, describes global business being conducted under unilateral EU rules even when other states continue to maintain their own rules \cite{bradford2012brussels}.
The proposed regulation on AI causing a de facto Brussels Effect is expected to be likely \cite{siegmann2022brussels}.
The European efforts are severe and according to \cite{floridi2019establishing}, it is evident that this will result in a legal framework based on the values of {\sl trustworthy AI} as laid out by the High-Level Expert Group (HLEG) on Artificial Intelligence, which aims to lay the foundation for the development of lawful, ethical, and robust AI systems  \cite{Ethicsguidelines2019,PolicyTAI2019}.

\subsubsection{Risk-based Regulation}
Fundamentally, the proposal classifies AI systems into three risk-based classes: {\sl Unacceptable risk}, {\sl high risk}, and {\sl low- or no risk}. See the Cap AI publication (\cite{CapAI2022}) for an informative summary. This risk-based approach has concrete consequences as AI systems classified as a potential source of {\sl unacceptable risk} will be prohibited. In contrast, for {\sl low or no risk} AI systems, no further action has to be taken.
However, a substantial part of AI systems already on the market or intended to be deployed in the future will likely be classified in the high-risk category, as every AI system referred to in Annex III of the European regulation shall be considered as high risk. Examples of criteria from Annex III are AI systems intended to be used
{\sl as safety components for recruitment or selection of natural persons to evaluate the creditworthiness of natural persons.}
Note that these examples represent only a small share of all affected AI systems. For further details including the full list in Annex III, see \cite{AIAct}.

\subsubsection{The Role of XAI in Regulation}
For all AI systems classified as high-risk, an extensive set of requirements is declared in Chap. II of the European regulation. These requirements can be relatively straightforward and directly touch upon transparency and explainability, among other points. Art 10 demands AI systems to ensure that they { \sl operate transparently and enable users to interpret the system's output appropriately} \citep{AIAct}, which is a direct call for XAI tooling in high-risk AI use cases.
However, regarding the concrete implementation, there are currently no standardized solutions but various approaches that may achieve these goals with XAI methods \citep{rai2020explainable,marques2022delivering}.
For example, XAI can improve model bias understanding and promotes fairness \citep{das2020opportunities} or increase transparency and detect adversarial examples \citep{kurakin2016adversarial}.

To sum up, there are concrete proposals for regulation, and a place is reserved for XAI to fulfill them. However, currently, we see only a few examples of how explainable AI can contribute to the development of trustworthy AI systems. Much future research remains to be done to enable XAI to be useful in certification and conformity assessments towards regulations, as the EC proposes. A result that is in line with the findings from our interviews

\section{XAI along the ML Life Cycle}\label{MLCycle}
\subsection{Data Collection and Understanding}
\subsubsection{Insights from the Interviews}

Collecting data for an industrial AI project is a complex task involving many roles, but the most important interaction is between data scientists and domain experts. The domain experts are expected to support data selection and the definition of the "right" data and play an essential role in supporting data scientists with data understanding and data quality evaluation. However, in industrial reality, data with the necessary variety, coverage, quantity, and quality is often scarce or restricted for internal or external reasons. 

While XAI typically plays its role during or after modeling, it was mentioned multiple times that XAI tools help in the iterated cycles between model building, data collection, and exchange with domain experts. Typical applications, where XAI is expected to support, in this context are investigating whether the available amounts and variety of data suffice and identifying issues with data quality, data bias, and wrong labels. One concrete approach mentioned is to detect relevant features via XAI methods and inspect the corresponding data quality more closely.  

\subsubsection{Academic Perspective on the Role of XAI for Data Collection and Quality} \label{subsec: data_academic}

The interviews revealed that the interaction with domain experts is important (which is extensively discussed from an academic perspective in section \ref{development}) and that 
 XAI is already recognized as a framework that can enhance data quality, collection, and understanding in practice. However, many relevant explainability techniques proposed by the academic literature do not seem prevalent. More specifically, a variety of methods aim to evaluate the influence of particular training data point on predictions, model performance, or the final model parameters. Note that such information can be utilized to address different challenges related to data quality and collection, like data valuation, noisy label detection, data subset selection, or guiding further data acquisition.

Typically, corresponding methods either utilize the knowledge created by an already trained model or evaluate models at different checkpoints during training. One line of work for this purpose uses the concept of Shapley values to identify the importance of individual training data points for the model performance \cite{ghorbani2019data, jia2019towards}, which can also be extended to entire training distributions \cite{ghorbani2020distributional, kwon2021efficient}. Since such approaches suffer from high computational costs, improving their scalability and effectiveness is an area of active research \cite{jia2019efficient,jia2021scalability,wang2021unified, wang2021improving}. 

Another way to identify influential data points for a given model is by considering influence functions \cite{cook1977detection,koh2017understanding}, which were initially designed to approximate the effects of Leave-One-Out (LOO) retraining. While multiple variations and adaptations have been proposed \cite{koh2019accuracy,khanna2019interpreting,basu2020second}, influence functions have successfully been applied to various tasks that can improve data quality, collection, and understanding. This includes training set subsampling \cite{wang2020less}, detecting memorized examples \cite{feldman2020neural}, interactive relabeling \cite{teso2021interactive}, resolving training set bias \cite{kong2021resolving} or to assist data augmentation \cite{lee2020learning}. Although these methods come with restrictive assumptions and have been demonstrated to be potentially unreliable \cite{basu2020influence}, recent results by \cite{bae2022if} suggest that they still might be useful in practice. If one has access to the training stage or intermediary model checkpoints, additional techniques can be utilized to identify influential or particularly difficult examples \cite{hara2019data,pruthi2020estimating, yeh2022first, agarwal2022estimating}.
A last category of methods explains predictions on test data based on the similarity to certain training data points \cite{kim2016critic, yeh2018representer, charpiat2019input,hanawa2021evaluation}. Understanding decisions via similarity can also increase data understanding through the lense of an AI model and can help to detect data set quality issues.

Overall, many XAI-related approaches exist that can help at the data collection and understanding stage based on an already trained model, but according to our interviews they are not yet popular among practitioners. Apart from that, some concepts related to XAI can also help increase data quality in the absence of any AI model. For example, in \citep{sebag2021shapley}, Shapley values are applied directly to database queries to identify causal tuples. How the interplay of AI and XAI can help at the level of databases to improve overall data quality is also an interesting direction for future work \citep{bertossi2020data}.

\subsection{Modelling} \label{development}
\subsubsection{Insights from the Interviews}
We found that developers of AI systems have a strong intrinsic need to understand their models and algorithms from a mathematical and algorithmic perspective. The majority of the interviewed data scientists reported that they want to understand how an AI model reaches its decisions, a fact that touches directly upon explainability. While the interviewees acknowledged that a broad magnitude of XAI methods for debugging, error tracing, etc., exists, lack of development support with XAI became apparent. Examples are, e.g., XAI-guided hyperparameter search, selection of network architectures, and model selection in general. However, in practice, by embedding XAI in the development and evaluation process, we have clear evidence that model analysis with XAI is not a one-off task but a continuous iterative process switching between development and evaluation. 

Besides the relevance of XAI for the developers themselves, we found a clear consensus that applied industrial AI development critically depends on collaboration with domain experts. This is important for selecting features, understanding model limitations, and incorporating domain knowledge into AI. Because the data alone often does not convey the domain expert's knowledge, data scientists are confronted with tremendous challenges if they have no access to domain knowledge, and it can take months to build up the necessary expertise. According to our findings, XAI can play its strengths for the cooperation between data scientists and domain experts. It bears significant advantages over classical approaches, such as communicating model results with descriptive statistics. However, these benefits come with the prerequisite that the XAI presents itself in the 'language' of the domain (expert). This means that explainability should be intuitive and must come in the semantics familiar to the domain expert.
\subsubsection{Academic perspective on the role of XAI for Modeling}
There already exist first approaches to support the development process with the help of XAI. In this section, we focus explicitly on the task of XAI-aided development, in contrast to XAI support for evaluation and testing, which forms the main body of research and is addressed in the following section.

One of the most crucial and also individual points during the development of an ML solution is finding appropriate hyperparameters for the model and application at hand. There are numerous efforts to automatize this process. However, the choice of a specific parameter and its effect is often opaque, even for highly experienced AI experts. Methods such as \cite{vanrijn2019hyper, moosbauer2021explaining} try to explain the influence of specific hyperparameters making the selection process more interpretable and lowering the level of required experience during the development phase. Also, model compression can be aided similarly by explainability techniques \cite{yeom2021pruning}.\\

Some interviewees further reported that the integration of domain expertise during the development process is regarded as an essential step. Multiple approaches exist trying to combine visual explanations generated by a model with prior knowledge \cite{rieger2020interpretations, ijcai2017p371}. This can offer an interpretable interface to formalize domain knowledge and also communicate the effect of the knowledge integration to domain experts. This concept has also been transferred to other data modalities such as text \cite{Liu2019IncorporatingPW}. A significant benefit of informing machine learning models with this kind of prior knowledge can be achieved by removing known spurious correlations in the training data set, which helps to increase the interpretability and robustness of the trained model \cite{ross2018improving}. 

However, preliminary to an extensive evaluation of a trained machine learning model, it is often unknown if spurious correlations are included in the data set to be used during training. 
Model explanations interpreted by human domain experts can indicate if a model has learned spurious correlations. Those might not be detected by data scientists, as mentioned in the interview responses, relying solely on (simple) performance metrics. Thus, in recently developed frameworks such as eXplainatory Interactive Learning (XIL) \cite{Teso2019ExplanatoryIM} and eXplainable Active Learning (XAL) \cite{ghai_xal}, the explanations of an intermediate state of the model are presented to domain or subject matter experts during an iterative training phase. This enables non-AI experts to argue with the model via an interpretable interface \cite{shao_towards_2020}. As with the previously described non-iterative methods, the form of this interface depends strongly on the task and, in particular, on the data format \cite{teso2022leveraging}. An interface consisting of rule-based explanations \cite{MLSYS2022_63dc7ed1} might be especially suited to tabular data, whereas visual explanations \cite{schramowski2020making} are more useful for image data. However, approaches such as those proposed by \cite{stammer2021right} also try to correct computer vision machine learning models by using rules based on high-level concepts and thus allow domain experts to argue on a semantic level they are used to.

\cite{ghai_xal} note that the continuous observation of the explanations of AI models during the training process can not only increase the trust in the system once it is deployed, but it also offers the opportunity for the non-AI-expert to calibrate their trust in the system during the learning process. This allows to estimate better in which situation the AI systems might fall short.

Moreover, also optimizing higher-level properties of explanations during training can help to increase the final model performance \cite{erion2021improving}.

\subsection{Evaluation}
\subsubsection{Insights from the Interviews}
Testing AI is reportedly a challenging task, and the non-availability of realistic operation environments was a major pain point for the interviewees. More generally, we learned that the quality of AI testing suffers from a lack of established test procedures. Currently, developers experiment with multiple approaches to test AI. One approach is black-box tests which care only about input-output relations. If possible, however, it was advocated to define edge cases in combination with the synthetic generation of test data or careful collection of test data in cross-functional teams, including domain experts. Especially the last point is strongly related to the vital importance of interaction with domain experts. Those need to set the baseline, help to confirm whether a model decides correctly, and define relevant test cases and scenarios. 

For those data scientists that advocate XAI for evaluation, we learned that they already make extensive use of state-of-the-art XAI methods to
globally debug models, trace down specific errors and conduct root cause reasoning. Among the named XAI tool stack are techniques such as Saliency Maps \cite{simonyan2013deep}, LIME \cite{ribeiro2016should} or GradCam \cite{selvaraju2017grad}.\footnote{References are given by the authors and were not part of the interview responses.} Many developers would welcome additional XAI methods if they are conveniently available open source and provide extra benefits for their domain. We also noted a need for unified interfaces that allow to combine the multitude of different XAI  methods. In contrast to these findings, we also received doubts regarding the technical feasibility of XAI methods to cope with complex models and critiques regarding the strict assumptions needed for many XAI methods. This helps to explain why some data scientists still place a focus solely on performance metrics and black-box testing.

Testing and validation are not only crucial tasks for ensuring the quality of AI but also touch upon the responsibility and liability that developers, product managers, product owners, or even whole companies take for their machine learning systems. Consequently, there is a need for risk control, ownership, and responsibility.

This is relevant because a central outcome from our focus group, the data scientists, is that they tend to dislike taking responsibility for their development because they struggle with accurately testing ML and the difficulties of understanding the risks associated with AI. The fact that XAI can support here was mentioned multiple times. The central part that XAI plays in this context is support for understanding the risks and potential failures associated with AI, identification of weak points, and partly relieving data scientists from their (perceived) burden of responsibilities.

\subsubsection{Academic perspective on the role of XAI for Evaluation}

From an academic perspective, the need for explainability for evaluation can be derived from many motivations. Many of them circle, in accordance with our findings from the interviews, centrally around the central problem of debugging and evaluating AI systems. 

For evaluation purposes, XAI can help on multiple levels. The most important distinction between the methods is commonly between {\sl intrinsically interpretable models} (sometimes called {\sl pre-hoc models} that are interpretable because of their simple structure or because of special architectural designs) and {\sl post-hoc methods} (that take a trained machine learning model as given). Especially for pre-hoc models, the separating line between modeling and evaluation becomes blurry.

Examples for {\sl pre-hoc models} are methods that allow being interpreted because of their relatively simple mathematical structure, such as linear regression, generalized additive models, and tree-based models \cite{molnar2020interpretable}. For this kind of model, the explanation coincides with the model itself, which can be a very convenient feature for the developer but restricts the choice of model to relatively simple ones. However, there also exists intrinsically interpretable components in more advanced models. For deep neural networks, leading examples are models with self-attention techniques, and especially transformers via attention maps \cite{chefer2021transformer} as well as concept-based approaches where latent representations can be interpreted in a meaningful way \cite{feifel2021leveraging}. 

{\sl Post-hoc methods} can be divided into {\sl model-agnostic methods} that do not require a specific model for explanations and {\sl model- or framework-specific methods} that are explicitly designed for a class or family of models. These approaches can again be subdivided into {\sl local and global techniques}. While local methods can support the developer in explaining AI models based on single instances, or data points, the global approach typically tries to reveal aspects or concepts that hold for the model as a whole. Leading examples for global methods in the context of neural networks are activation maximization approaches \cite{nguyen2016synthesizing}, concept-based approaches \cite{kim2018interpretability} and knowledge extraction methods \cite{yang2018global, frosst2017distilling}. Popular local methods for neural networks are feature attribution methods such as Saliency Maps \cite{simonyan2013deep}, Layer-wise Relevance Propagation (LRP) \cite{bach2015pixel}, and information-theoretic approaches \cite{chen2018learning}. Leading examples for model-agnostic perturbation-based approaches include, in particular, local interpretable model-agnostic explanations (LIME) \cite{ribeiro2016should} and Shapley additive explanations (SHAP) \cite{lundberg2017unified}. Furthermore, a whole literature on the identification of influential instances \cite{koh2017understanding}, and adversarial examples \cite{goodfellow2014explaining, su2019one} exists.

The doubts expressed by some interview partners concerning the fidelity and reliability of the generated explanations are also an increasingly debated issue in academia. The lack of evaluation metrics to quantify the fidelity of different explanation methods makes it difficult even for data scientists to estimate which method to use for a particular application. \cite{krishna2022disagreement} describing the {\sl disagreement problem}, demonstrating that different XAI methods can generate contradicting explanations for a given prediction, underpins this problem. To tackle the issue \cite{rudin2019stop} proposes to use intrinsically interpretable pre-hoc models instead of post-hoc explanations for black-box models. On the other hand, efforts such as by \cite{bodria2021benchmarking} establish a holistic topology and comprehensive benchmarking to ease the choice of applying XAI methods in practice. With this in mind, explaining something depends not only on the transmitter but also on the interpretation and, therefrom, drawn conclusions of the human recipient. However, there is currently only a little work trying to take the latter part into consideration \cite{yang2022psychological}.

\subsection{Deployment and Productivization}
\subsubsection{Insights from the Interviews}

Regarding deployment, our interview partners strongly emphasized the inclusion of end users. At best, AI systems should be designed to integrate seamlessly into existing systems without too many changes compared to current systems that should be replaced or enhanced. We also gathered evidence that acceptance of the new ML system increases if the ML system provides guidance for the user, drill-down possibilities, and root-cause reasoning. Hence, although the AI should often be concealed to the end user, so that the user does not percieves the black box so obviously, it is highly useful to use XAI functionalities to enable user-friendly systems that do not appear as black-box AI. This applies mainly to AI agnostic end users whoe are not interested in the model details.

Generally, it was noted multiple times that the customer and users need to understand the model before deployment. A requirement that strengthens if the end users have some responsibility which increases their need to explain why an error or decision occurred in their system. However, this raises the need for building suitable XAI interfaces. Additionally, we received many recommendations on what measures should be taken to ensure that the XAI comes in the required form. Motivations for these recommendations are that currently, it is often unclear how XAI should be used and which level of granularity is needed for the user.
First and foremost, the importance of integrating the user in the design of XAI systems was highlighted. This includes user studies, user tests, user feedback, and carefully designed user experience (UX) for the interfaces. Hence, depending on the user and the use case, it is of vital importance that XAI comes in the form and semantics a user is used to, to bring real benefit.

Another issue of tremendous importance for deployment is whether current XAI methods scale for actual deployment. We gathered many insights showing that scaling XAI comes with multiple challenges that are not only of technical nature. First, many interview partners highlighted that XAI is computationally complex, which hinders XAI in deployment. Furthermore, it was noted that XAI does not scale to other data structures besides images. Another reason that prevents scaling is the lack of experts to interpret XAI methods' outcomes. On the other hand, we gained the insight that XAI itself can be an enabler of scaling ML if it enables non-AI experts to build, understand and use AI models. 

\subsubsection{Academic perspective on the role of XAI for deployment and scaling of XAI}
Current research is in line with our findings insofar that, e.g., \cite{adadi2018peeking} highlights that academic research in XAI is strongly biased towards algorithmic improvements but lacks the human aspects. However, the central finding above is that XAI can only enfold its full strength if the explanations are presented in the 'language' of the domain experts and end users. To be precise, this can mean that the explanations need to undergo a semantic transformation (see, e.g., \cite{de2020human}) such that a model is explained in another semantic domain as it was trained in. An industrial example could be, e.g., a model is trained on raw time series data but explained in the frequency domain. Alternatively, more generally, in the domain, a user feels comfortable instead of simply taking the one it was trained on.
\\
{\sl The role of human-machine interaction and visual analytics}\\
Independently of the semantics, visualization has a crucial role to play in the communication of XAI. However, using state-of-the-art visualization techniques, often shipped with publications and packages for local and global XAI, such as waterfall, force, or bar chart feature importance plots, is often insufficient to achieve this goal. The reason is that the methods and corresponding visualizations are designed generically for data scientists and AI researchers from various domains instead of non-AI experts. Hence, a presentation of explanations tailored to the end user's mental model of the domain is needed to ensure highly effective workflows that enable seamless and intuitive insight-gaining. 
This conforms with the principles of the Visual Analytics (VA) design and implementation process that aims to create the most efficient, expressive, and appropriate visualization methods by taking the major factors of the design triangle into account: data, users, and tasks \cite{Miksch13amatter}.

According to the nested model of \cite{5290695}, the first level of visualization design characterizes the problems, tasks, and data for the target users. Following this model, highly specific XAI visualizations need to be implemented for each particular domain, user, and data combination. Consequently, a scalability problem arises since new effort must be spent for each new combination. 
However, from another perspective, there is hope that common VA principles, best practices, and theories will "provide economies of scale" \cite{Keim2008}. However, those can only be reached at higher, more general levels and therefore do not solve the scalability issues as discussed above.

Commercial off-the-shelf visualization software such as Tableau or Power BI  offers this scalability with powerful, reusable visualization techniques that can be combined to create individual dashboards. However, domain experts usually are not willing or able to create their own complex dashboards to visualize the XAI output as discussed in  \cite{PB-VRVis-2017-019}. Furthermore, much XAI expertise is needed anyhow to create meaningful output from the results of these methods.

Recently the combination of VA and XAI, Visual-based XAI (vXAI), is picking up more and more traction in the research community. 
However, until now, no common, standardized visual approaches to present local or global explanations for XAI methods for different types of data, models, and domains emerged \cite{ALICIOGLU2022502}. Consequently, it remains an open research challenge whether the generalization of domain-specific vXAI solutions can solve the inherent scalability problem.
\\
{\sl XAI for Deployment and Scaling of XAI}
\\
The problem of poor scalability of many XAI methods was mentioned multiple times in the interviews and indeed constitutes a significant restriction in practice. While gradient-based methods applicable to differentiable models typically can be computed fairly efficiently on modern GPUs \cite{ancona2019gradient}, scalability is particularly problematic for model-agnostic techniques that require a multitude of model evaluations. A prominent example is, for instance, Shapley values which also belong to the most popular approaches among practitioners \cite{bhatt2020explainable}. In \cite{van2022tractability}, the authors show that the computation of Shapley values is intractable even for simple, commonly used models such as logistic regression. Therefore, appropriate computation and approximation strategies have been proposed in the academic literature \cite{lundberg2017unified, castro2009polynomial, covert2021improving}. Moreover, knowledge about a particular graph structure within the data can also be leveraged to speed up computation \cite{chen2018lshapley}. Model-specific versions also exist to increase the efficiency for deep neural networks further \cite{lundberg2017unified, ancona2019explaining,wang2022accelerating} or for tree ensembles\cite{lundberg2018consistent, yang2021fast, mitchell2022gputreeshap, yu2022linear}.
Another way to address scalability is by considering XAI methods that learn a separate model to create explanations \cite{chen2018learning, yoon2018invase, jethani2021have}. Once trained, such an explanation model allows to retrieve explanations quickly at inference time, enabling better scalability during deployment. This idea can also be used to learn the estimation of Shapley values explicitly \cite{jethani2021fastshap, covert2022learning}. Nevertheless, more research and implementation efforts are needed to increase the overall scalability of different XAI methods in general via more efficient computations or hardware utilization. 

\subsection{Monitoring and Maintenance} 
\begin{table}[]  \footnotesize
	\caption{Requirements towards XAI systems that support monitoring and maintenance}
	\begin{tabular}{l|l}  \hline \hline
		Enablement for monitoring& XAI needs to support communication with domain experts.                                                                \\
		& XAI should enable an understanding of the model functionality.                                                  \\
		& XAI should enable an understanding of the decisions of the model.                                                   \\ \hline
		Incident handling support for monitoring            & Providing root cause analysis for errors. \\
		& Indicating the type of error/incident.                                                 \\
		& Indicating the criticality of errors/incidents and their urgency.                                                    \\
		& Providing recommended actions.                                                                                     \\
		& Supporting with what-if scenarios if an action is not initiated.                                                   \\ \hline
		Support for ad hoc questions/requests                              & Explaining why a certain result/decision is given.                                                                 \\
		& Enabling drill-down possibilities.                                                                                 \\
		& Representing the certainty of the results/decisions.                                                               \\ \hline
		Support for maintenance                                                 & Highlighting important features.                                                                                   \\
		& Identifying distribution shifts.                                                                                   \\
		& Suggesting how to adjust the model.      \\ \hline \hline                            
	\end{tabular}
	 \label{requirements}
\end{table}
\subsubsection{Insights from the Interviews}
We received very diverse feedback on who is currently, or will be, responsible for monitoring and maintaining ML systems. The following roles of non-AI experts have been named: Application and automation engineers, service technicians, operators, or even a new job profile (e.g., in analogy to DevOps engineers). Since these roles are, per se, not necessarily equipped with ML expertise, the persons filling the roles will need training, well-designed interfaces, dashboards, and further tooling to do their job.

In contrast to the listed persons above, it was often noted that non-experts could not master the range of tasks and responsibilities needed for monitoring and maintenance without profound AI knowledge, shifting the role again towards AI engineers and data scientists. The main reason for placing this role for monitoring and maintenance is their competencies since monitoring and maintaining AI comprises many tasks only data scientists can do. It was also mentioned that the tasks associated with monitoring and maintaining ML are too broad for a single person, and instead, whole teams will be involved. This brings the advantage that teams do not depend on the knowledge and skills of one person and enables specialization. E.g., there will likely be a split between teams because the job profiles and needed skills for monitoring and maintenance differ strongly. However, an obvious obstacle to this proposal is the current lack of skilled and educated AI experts.

For the question of how to concretely monitor AI systems, no standard solutions exist. On the contrary, there are fundamentally distinct views on the topic. Some persons recommend monitoring the functionality of a model, while on the other hand, it is also argued that monitoring is rather about the data than the model. We even gained the insight that for some use cases, neither the model nor the data should be monitored but a higher-level business metric, such as throughput. This would imply that AI investigation and maintenance are only conditional to the failure of the higher-level business metric.

We found that the relevance of XAI is especially pronounced for supporting running AI systems and monitoring tasks. The reasons are manifold, but the key arguments are AI maintainability, keeping trust high, and safety concerns for critical systems/decisions.  

Regarding the required form of XAI in monitoring and maintenance, we received various requirements and examples summarized in Table \ref{requirements}. Centrally, we learned that any person who needs to monitor AI systems needs to be enabled when it comes to an understanding of the model to monitor and efficient communication with domain experts on the one hand and AI developers on the other hand. Furthermore, incident handling is a non-trivial task that needs to be supported with tooling that enables the analysis of incidents with XAI as well as semi-automated action initiation. Furthermore, persons that monitor need to be able to give ad hoc answers to requests. Again, XAI can hereby play a vital role in explaining results, showing drill-downs, and demonstrating uncertainty measures. Lastly, the maintenance of AI models relies on knowing which features are important in what way and suggestions on how the model should be adjusted after a breakdown of the AI system.
\subsubsection{Academic Perspective on Monitoring/Maintanence}
Even though the link between monitoring ML models and explaining their predictions is quite compelling, which is also reflected in the interviews, there are only a few dedicated approaches in academia working on that intersection. The existing methods can be split by their intention of using explainability methods during the monitoring process.\newline

The authors in \cite{sagemaker_monitor} use the shift in the explanations of a model as an early indicator for a potential performance degradation. Furthermore, they claim that by monitoring the attention and maintenance, ML-model attributes to certain features can be used to identify emerging biases in its predictions with respect to fairness requirements. Similarly, \cite{mougan2022explanation} evaluates the connection between the change of a model's explanations and its performance.\newline
On the other hand, \cite{pmlr-v130-budhathoki21a} is trying to explain a detected decline in performance by providing actionable insights about its cause. Rather than directly explaining the decline in performance, \cite{zhang2022why} use XAI methods to investigate a complex data drift. Both approaches require a causal graph of the underlying data generation process. The authors in \cite{mougan2022monitoring} claim to attribute model deterioration to individual features without needing a causal graph of the data generation process or labels for incoming data. Explanations hinting at the reason for the degrading model performance can be utilized to mitigate the cause of a performance drop and facilitate the communication with domain experts to reestablish a reliable predictive system jointly.

\section{Findings regarding intital hypotheses} \label{hypotheses_summary}
Since our research and interviews was guided by the hypotheses outlined in Table \ref{hypotheses} we provie a summary of our central learnings with respect to these hypotheses below.
\subsection{Hypotheses Data Scientists}
\subsubsection{XAI support the communication with domain experts} is a hypothesis we can confirm. We indeed gathered much evidence that interaction with domain experts is vital, especially for data understanding, modeling, and testing, and is already in practice supplemented with XAI methods. However, while the potential is clear, many data scientists would like more tools to improve on this point.
 
\subsubsection{XAI improves the development process} also is supported by our interviews. The main issue here is, nevertheless, that XAI primarily supports for evaluation of models but is not recognized as being helpful for the modeling part itself by practitioners.

\subsubsection{XAI improves AI testing} may appear to be already answered by our notes on the previous hypothesis. However, although, in theory, a considerable research body for evaluation with XAI exists in the real-world industries, testing often has another connotation. I.e., while data scientists agree that XAI methods help to evaluate whether a model captures meaningful relationships, it often fails to support them with the critical decision of whether the model is really ready for deployment. A question that also invokes challenges such as model behavior on out-of-distribution data, feature drift, etc., which cannot be solved solely with XAI.

\subsection{Hypotheses Monitoring}
\subsubsection{XAI supports the task of monitoring} We gathered much evidence that explainability will be very relevant for the task of monitoring AI. Mainly because proper monitoring potentially requires functinalities such as root cause reasoning, etc. Furthermore, we learned from the interviews that the task of monitoring will be potentially be filled by diverse roles which do not necesarily come with a data science and machine learning background and hence need support in unerstanding the reasoning of AI systems.
\subsubsection{XAI can support the task of maintaining AI} With respect to maintaining, we also learned that XAI will be relevant in order to highlight which features are relevant, e.g., for declining performance but also to guide the persons that need to mainten the model how to do this most efficiently.
\subsubsection{XAI can support root cause analysis, commissioning, and other tasks} Similar as above we found mcuh evidence that these tasks could strongly benefit from XAI tooling.

\subsubsection{XAI can support audits} Here, we cannot report too much evidence because we found that among the interviewees it is still unclear to which extent there will be regulation, standardization or regular audits for industrial AI.

\subsection{Hypotheses Business}
\subsubsection{AI and XAI are among the strategic priorities} While AI is definitvely among the strategic priorities of many industrial companies it is harder to confirm this for XAI. This is mainly because from a business or management perspective trustworthy AI is more than a technical issue but also related to bussiness habits and other factors such as branding.

\subsubsection{XAI bridges gaps in cross-functional teams} Here, we can at least confirm the strong need for bridging gaps between data scientists, domain experts and end users at various stages of the life cycle togehther with evidence that XAI lowers barriers for interactions between those roles.
\subsubsection{XAI is needed as a distinguishing factor (from competitors)} This hypotheses can be confirmed in the sense that we learned that transparency and explanability rank high among the features demanded by the AI market.

\section{Conclusion} \label{conclusion}
In this paper, we represent the findings of 36 interviews regarding the relevance of XAI in an industral context in contrast with the current state-of-the-art academic research. We found that XAI already plays a vital role at various stages of the AI life cycle and is expected to grow in importance. Furthermore, we found that tangible business requirements such as the need for disginguishing factors and potentially upcoming regulations are a driver for XAI in the industries.

 The interviews and our literature research also allows us to confirm the findings of previous studies. It still holds that most attention of academia is on the data scientists and their main task of iterating between model development and evluation. However we also found that there already exists a non-negible body of literature that tries to address other stakeholders and stages of the ML life cycle. 
 
 Lastly, we also identified the need for more resarch along the ML life cycle being a demand of our interview partners. This holds especially for the interplay between XAI, AI, data collection and cleansing as well as XAI enhanced monitoring and maintenance of AI models. Generally, we see two central mismatches as central outcomes of our research. On the one hand, is appears as if the academic XAI toolbox is not yet fully utilized in practice. On the other hand, practioneers demand techniques and tools that do not yet exist. Hence, our findings can be interpreted as a call for practitioneers to widen their view on the available methods but also  for directing more research effort into enabling explainability for different stakeholders such that XAI can unfold its full potential wherever needed at the ML life cycle.

\bibliographystyle{splncs04}   
\bibliography{literature} 

\appendix

\end{document}